\documentclass[letterpaper, 10 pt, conference]{ieeeconf}  
\IEEEoverridecommandlockouts                              
\usepackage{amsmath}
\usepackage{amssymb}
\usepackage[breaklinks=true,bookmarks=false]{hyperref}
\hypersetup{
    colorlinks=true,
    linkcolor=blue,
    filecolor=magenta,      
    urlcolor=cyan,
}
\usepackage{cite}
\usepackage{multirow}
\usepackage{subcaption}
\usepackage[pdftex]{graphicx}
\usepackage[]{algorithm2e}
\usepackage{lipsum}
\usepackage{enumerate}
\usepackage{tabularx}
\usepackage{color}
\usepackage[english]{babel}
\usepackage{blindtext}
\usepackage{authblk}

\usepackage{array}
\newcolumntype{L}[1]{>{\raggedright\let\newline\\\arraybackslash\hspace{0pt}}m{#1}}
\newcolumntype{C}[1]{>{\centering\let\newline\\\arraybackslash\hspace{0pt}}m{#1}}
\newcolumntype{R}[1]{>{\raggedleft\let\newline\\\arraybackslash\hspace{0pt}}m{#1}}

\newcommand\blfootnote[1]{%
  \begingroup
  \renewcommand\thefootnote{}\footnote{#1}%
  \addtocounter{footnote}{-1}%
  \endgroup
}

\begin{document}

\title{\LARGE \bf
A Multimodal, Full-Surround Vehicular Testbed for Naturalistic Studies and Benchmarking: Design, Calibration and Deployment}


\author[1]{Akshay Rangesh}
\author[1]{Kevan Yuen}
\author[1]{Ravi Kumar Satzoda}
\author[1]{Rakesh Nattoji Rajaram}
\author[2]{\\Pujitha Gunaratne}
\author[1]{Mohan M. Trivedi}
\affil[1]{Laboratory for Intelligent and Safe Automobiles (LISA), UC San Diego}
\affil[2]{Toyota Collaborative Safety Research Center (CSRC)}



\maketitle
\thispagestyle{empty}
\pagestyle{empty}

\blfootnote{$^\dagger$\href{https://drive.google.com/open?id=1CsNyxxmHABsL0CpMtIkiF8r2S2HrKo7m}{Dataset download}}


\begin{abstract}

Recent progress in autonomous and semi-autonomous driving has been made possible in part through an assortment of sensors that provide the intelligent agent with an enhanced perception of its surroundings. It has been clear for quite some while now that for intelligent vehicles to function effectively in all situations and conditions, a fusion of different sensor technologies is essential. Consequently, the availability of synchronized multi-sensory data streams are necessary to promote the development of fusion based algorithms for low, mid and high level semantic tasks. In this paper, we provide a comprehensive description of LISA-A: our heavily sensorized, full-surround testbed capable of providing high quality data from a slew of synchronized and calibrated sensors such as cameras, LIDARs, radars, and the IMU/GPS. The vehicle has recorded over 100 hours of real world data for a very diverse set of weather, traffic and daylight conditions. All captured data is accurately calibrated and synchronized using timestamps, and stored safely in high performance servers mounted inside the vehicle itself. Details on the testbed instrumentation, sensor layout, sensor outputs, calibration and synchronization are described in this paper.

\end{abstract}
\section{Introduction}

\begin{figure}[!t]
  \centering
  \includegraphics[width=1\linewidth]{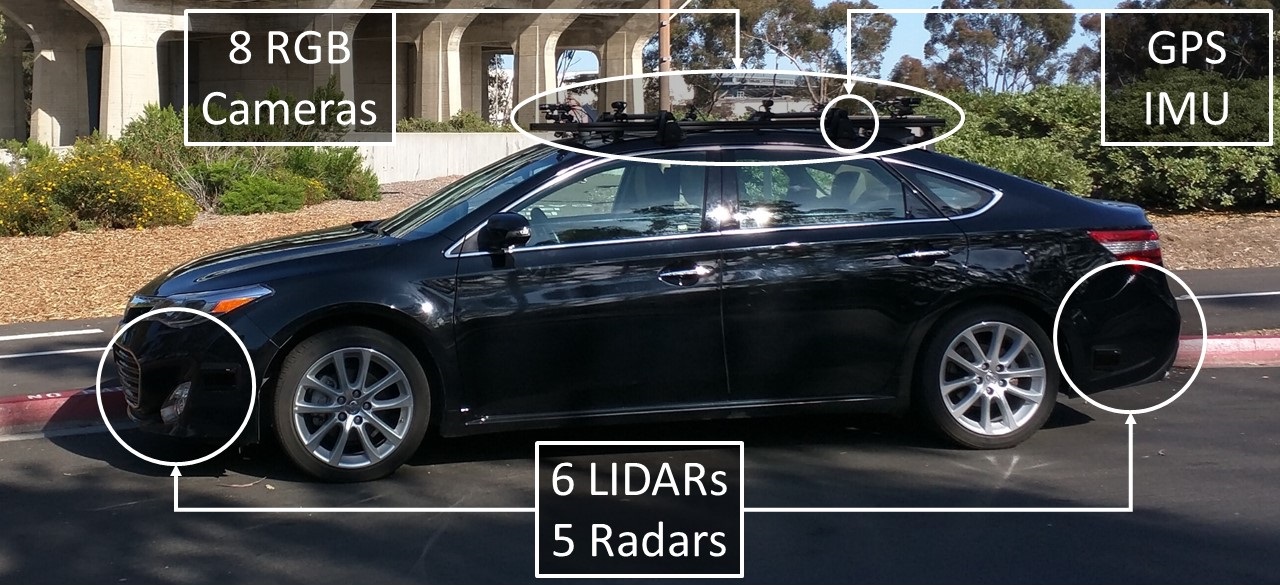}
  \caption{\textbf{LISA-A:} A Toyota Avalon equipped with eight RGB video cameras, 6 LIDARs, 5 radars, and a GPS/IMU unit.}
  \vspace{-5mm}
  \label{fig:002_car_image}
\end{figure}

The proposed testbed LISA-A (Fig.~\ref{fig:002_car_image}) equipped with a suite of active and passive sensors has been driven on the roads of San Diego, California for over half a year. This testbed was created with the primary goal of enabling algorithms that make use of multi-sensory inputs to carry out tasks essential to driver assistance and autonomy. An intelligent moving platform generally consists of \textit{proprioceptive} and \textit{exteroceptive} sensors. The former is responsible for sensing the state of the ego-vehicle, while the latter is used to sense the ambient surroundings. In order to offer researchers complete flexibility while creating fusion based algorithms using exteroceptive sensors, our testbed offers \textit{full surround} coverage from cameras, radars and LIDARs. This provides a more holistic picture of the environment during the development of algorithms for a variety of tasks.

The rest of this paper is organized as follows: In section \ref{related}, we list some previous studies on existing testbeds, and contrast their approach with ours. Section \ref{testbed} describes the sensor layout that we chose to permit full-surround coverage, and section \ref{sensor} provides details about the output of each sensor. Section \ref{calibration} explains how the sensors were calibrated, and how time synchronization between them is achieved. Finally, we discuss future avenues for research and summarize our work in section \ref{conclusion}.

\begin{table*}
\centering
\caption{Comparing our testbed with related vehicular testbeds and published datasets.}
\label{tab:relres}
\resizebox{15cm}{!}{
\begin{tabular}{lllllllllllll}
      & \rotatebox{90}{Year}  & \rotatebox{90}{Camera Resolution} & \rotatebox{90}{Camera FPS }  & \rotatebox{90}{Total Cameras} & \rotatebox{90}{Stereo Rig} & \rotatebox{90}{Panoramic Camera Array} & \rotatebox{90}{LiDAR} & \rotatebox{90}{Radar} & \rotatebox{90}{GPS/IMU} & \rotatebox{90}{Vehicle Parameters} & \rotatebox{90}{Long Term Sequences} & \rotatebox{90}{Scene Details} \\

\hline

Ford Campus \cite{pandey2011ford} & 2011  & 800x600  & 8 Hz  & 6   &            &  \checkmark     &  \checkmark     &       &  \checkmark     &       &       & US/Urban \\

TME \cite{TMEMotorwayDataset} & 2012  & 1024x768 & 20 Hz   & 2     &  \checkmark      &       &  \checkmark     &       &       &       &  \checkmark     & Italian/Hwy \\

Annieway \cite{KITTI2012} & 2012  & 1382x512 & 10 Hz  & 4   &  \checkmark         &       &  \checkmark     &       &  \checkmark     &       &       & German Urban/Hwy \\

LISA-Audi \cite{lisaaudi} & 2012  & 1024x522 & 25 Hz  & 1  &            &       &  \checkmark     &  \checkmark     &  \checkmark     &  \checkmark     &  \checkmark     & US/Hwy \\

Cityscapes \cite{Cordts2016Cityscapes} & 2016  & 2048x1024 & 17 Hz& 2    &  \checkmark         &       &       &       &  \checkmark     &  \checkmark     &       & European/Urban \\

LISA-Surround \cite{MJ2016} & 2016  & 2704x1440 & 12 Hz  & 4 &            & \checkmark     &       &       &       &       & \checkmark    & US/Hwy \\

\textbf{LISA-A} & 2017  & 1600x1200 & 30 Hz  & 8  &           & \checkmark     & \checkmark     & \checkmark     & \checkmark     & \checkmark    & \checkmark     & US/Hwy \\
\hline
\end{tabular}%

}
\end{table*}

\section{Related Vehicular Testbeds} \label{related}

Table \ref{tab:relres} provides a detailed comparison of the testbed described in this work with related testbeds and published datasets. Evidently, a majority of the testbeds are constructed using a subset of the extensive instrumentation employed in this work. In particular, our work stands apart in the number of cameras employed for full panoramic capture at 30 Hz. Most testbeds with panoramic capture employ a single mount with small baseline distances such as a Ladybug camera \cite{pandey2011ford}, yet the capture of high resolution surround data with 8 cameras is significantly more challenging and insightful. In particular, employing 8 distributed cameras provides an unobstructed view of the scene with little to no distortion. The testbed in \cite{MJ2016} employs four wide angle cameras for full surround image capture, but they lack the rest of the instrumentation required for safe, multi-modal autonomous navigation and planning. Table \ref{tab:relres} also emphasizes how most existing datasets involve only short sequences (e.g. KITTI \cite{KITTI2012}), which cannot be used to carry out analyses for behavior and planning-level operations. On the other hand, we have tailored our testbed for this very task, as multi-modal full surround data (from cameras, radars, and LIDARs) can provide essential cues for long-term behavior monitoring and planning.

It must also be pointed out that our most similar testbed is detailed in \cite{lisaaudi}, with the remaining testbeds capturing data in non U.S. settings. This is critical for development of autonomous vehicles in the U.S., which presents a different set of challenges than some of the European roads and landscapes, including different navigational norms, road markings, road weather conditions, etc. 

\section{Testbed Design and Architecture} \label{testbed}
The selection of sensor modalities and their placement is crucial for highly automated vehicles. Vehicles with a high level of autonomy require extensive surround instrumentation of multiple sensor modalities. Our unique testbed incorporates LIDARs, radars, cameras, and GPS/IMU sensors to holistically observe and understand the scene. We list the entire suite of sensors in our testbed below. All sensors have been mounted aesthetically to support naturalistic data collection.

\begin{itemize}
  \item 8 $\times$ Point Grey Flea3 RGB Cameras
  \item 6 $\times$ Kowa Lens with 4.5mm Focal Length
  \item 2 $\times$ Kowa Lens with 3.5mm Focal Length
  \item 6 $\times$ ibeo LIDARS (ibeo LUX Fusion System)
  \item 4 $\times$ Delphi SRR2 Radars
  \item 1 $\times$ Delphi ESR Radar
  \item 1 $\times$ Mobileye Driver Assistance System
  \item 1 $\times$ IMU/GPS
\end{itemize}

The 8 Point Grey cameras are interfaced using an RJ45 ethernet connection split across two server computers installed with 64-bit Ubuntu 14.04 in order to handle the data capture and compression load. Each server runs 2 Intel Xeon Processors E5-2630 v3 with a total of 32 threads. The two cameras on either side of the vehicle are fitted with lenses with smaller focal lengths to provide a wider field of view capable of completely capturing vehicles directly next to the ego-vehicle. The remaining six cameras use a slightly higher focal length in order to retain more details on distant vehicles, while the wide angle lenses ensure sufficient visual overlap between neighboring cameras so that there are no blind spots.

\begin{figure}[!t]
  \centering
  \includegraphics[width=1\linewidth]{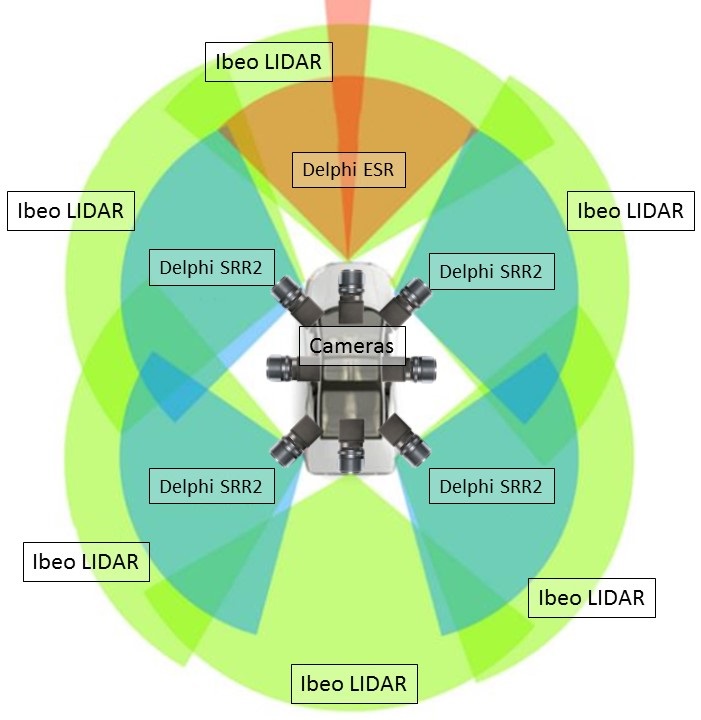}
  \caption{\textbf{Sensor layout:} Our Toyota Avalon equipped with 8 RGB video cameras, 6 LIDARs and 5 radars. Approximate range and field of each sensor is shown in color.}
  \vspace{-5mm}
  \label{fig:002_sensor_layout}
\end{figure}

We equipped the testbed with six Ibeo LUX laser scanners and Delphi radar sensors to capture the surroundings of the ego vehicle. There are 3 LIDARs each installed in the front bumper area and the rear bumper area to establish a full surround view using a total of 6 LIDAR sensors. Two side-looking radars are installed on the front and rear bumper areas, along with a long-range front-looking radar to detect vehicles farther along the road, totaling to five radars in all. A top-down view of the sensor layout is shown in Fig. \ref{fig:002_sensor_layout}. In addition to these surround sensors, the Mobileye Driver Assistance System is also installed underneath the center rearview mirror on the front windshield. This provides collision warnings and lane departure events in addition to state-of-the-art lane information. Vehicle dynamics are also recorded from the vehicle's Controller Area Network (CAN bus) and the IMU/GPS devices. The sensors are meant to complement each other, so that fusion of information across modalities can provide a more comprehensive scene analysis.

\section{Sensor Output} \label{sensor}

\begin{figure*}[!t]
  \centering
  \begin{tabular}{c}
        \includegraphics[width =0.95\textwidth]{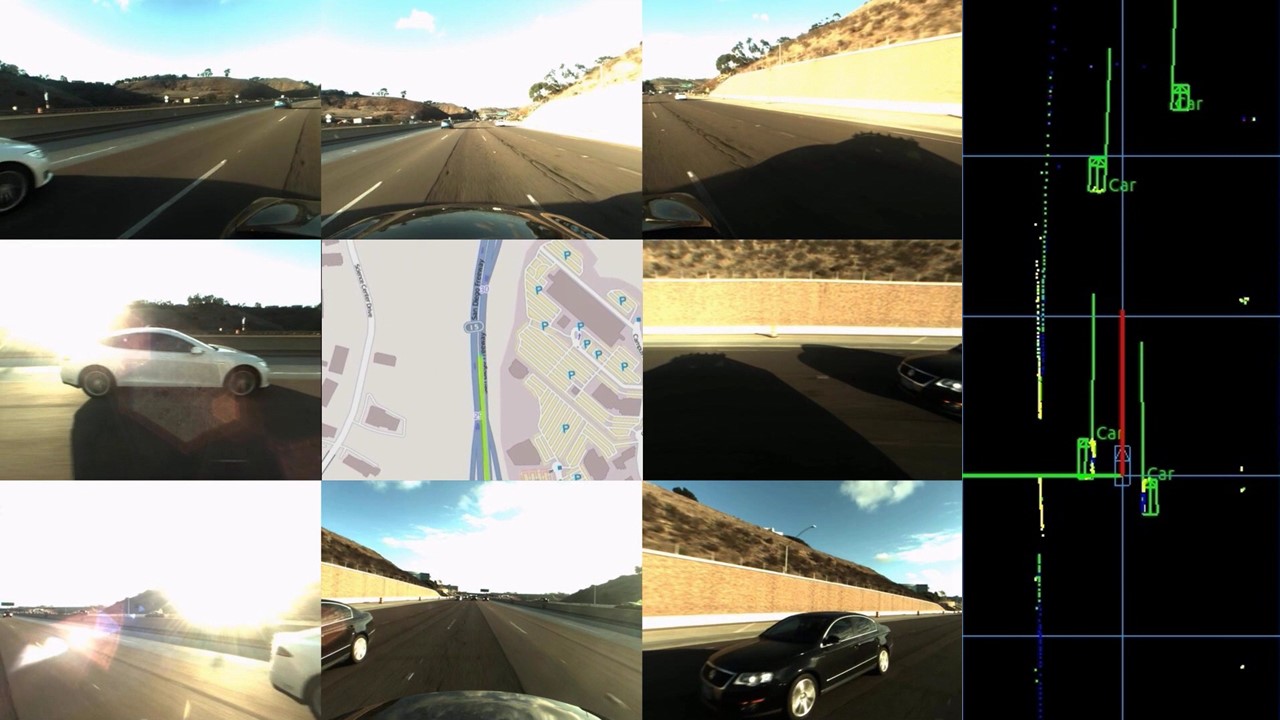}
  \end{tabular}
  \caption{\textbf{Camera, GPS, and Lidar output:} Example output from 8 (calibrated) cameras along with the GPS location visualized on a map (center). Top-down view of full surround LIDAR points and object list (right).}
  \vspace{-5mm}
  \label{fig:003_surround_view_maps}
\end{figure*}

\textbf{Cameras:} The 8 Point Grey cameras record at 30 FPS with a resolution of $1600\times1200\times3$ using Point Grey's Fly Capture API library, and the videos are compressed using the XVID codec. An example output is shown in Fig. \ref{fig:003_surround_view_maps}. The remaining sensors record data by using the PolySync API to retrieve information from the PolySync bus stream.

\textbf{LIDARs:} The four-layered ibeo LUX scanners (LIDARs) output LIDAR point clouds in the $(x,y,z)$ format and can classify and track objects up until 200 meters with a horizontal field of view of 110 degrees at 12.5 Hz. The LIDARs are used to reliably detect all obstacles in the scene, such as cars, trucks, pedestrians, cyclists, as well as other road obstacles.

\textbf{RADARs:} The four side SRR2 radars provide blind spot detection and monitoring of vehicles adjacent to the ego-vehicle. This is essential for recognizing overtaking vehicles and planning lane change maneuvers. The forward Delphi ESR radar provides wide coverage at mid-range and high-resolution long range coverage. This allows vehicles who cut in from adjacent lanes to be detected. At mid-range ($\sim$60m), the radar provides 45 degrees field of view, and at long-range (up to 174 m) 10 degrees. Data is acquired at 76 GHz.

\textbf{Mobileye Sensor:} This sensor provides semantic scene cues, including an accurate lane detection system which can be used to localize the ego-vehicle in the ego-lane. This provides the following information for both the left and right lane markers -
\begin{itemize}
\item Lane marker type
\item Lane marker width
\item Lane marker type
\item Lane marker width
\item Lane marker heading angle
\item Lane marker view range
\item Lane marker offset
\item Lane curvature
\end{itemize}

\begin{figure}[!t]
  \centering
  \includegraphics[width=1\linewidth]{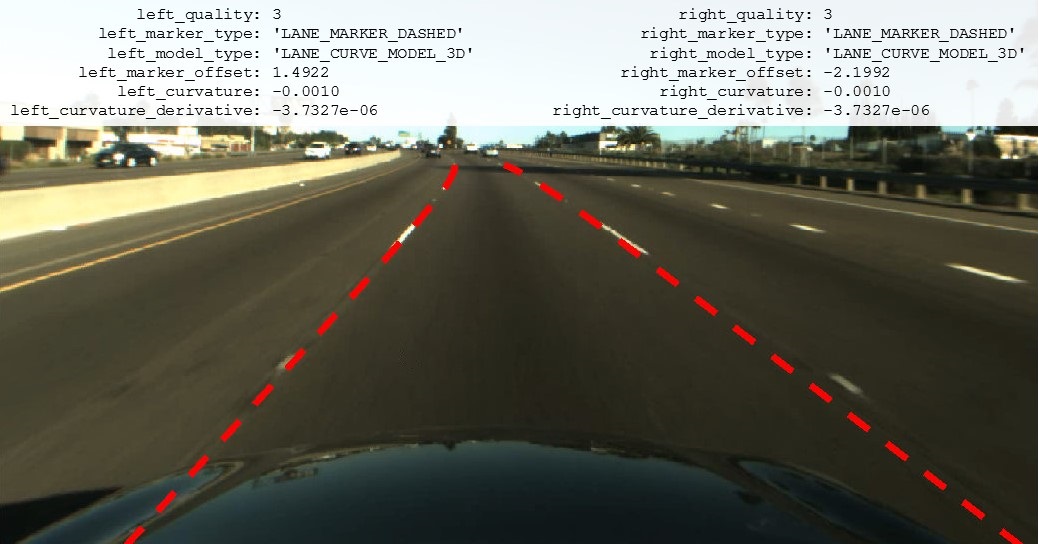}
  \caption{\textbf{Mobileye output:} Ego-lanes projected onto the front camera image, along with lane model information.}
  \vspace{-5mm}
  \label{fig:003_mobileye}
\end{figure}

With this setup, we are now able to not only localize the ego-vehicle accurately within each lane, but also measure lane deviation, lane changes and localize surround vehicles to lie within a particular lane (which would be very useful for threat assessment). Fig. \ref{fig:003_mobileye} shows a visualization of the left and right lane markers projected onto the front camera image, based on the models described above.

\begin{figure}[!t]
  \centering
  \includegraphics[width=1\linewidth]{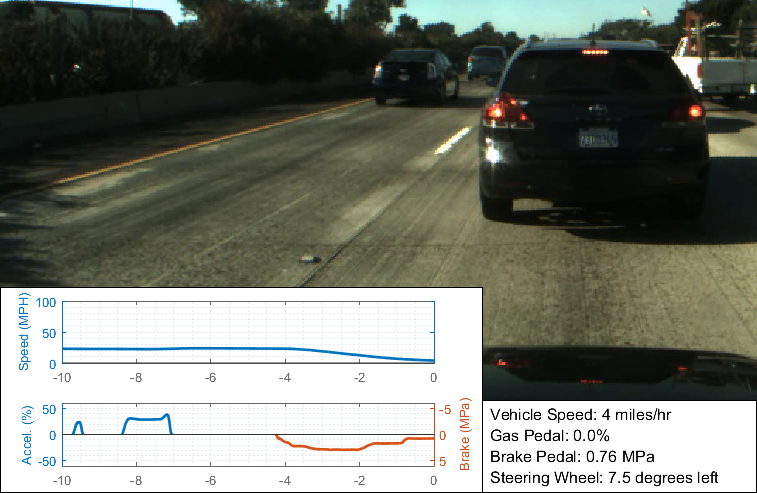}
  \caption{\textbf{CAN output:} A subset of the CAN data fields being shown along with front-facing camera image (cropped): vehicle speed, gas/brake pedal pressures and steering wheel angle.}
  \vspace{-5mm}
  \label{fig:003_can}
\end{figure}

\textbf{GPS and Vehicle Dynamic Sensors:} An Xsens MTi G-710 sensor provides high quality GPS, gyro, and accelerometer measurements. The data is useful for accurate ego-vehicle localization, and for leveraging map information (e.g. scene type, lanes, and landmarks). OpenStreetMaps is one such tool that provides rich annotations for every region along with rendering tools to visualize map in a manner similar to that available on other commercial software, all of which can be accessed by providing suitable GPS coordinates as recorded by our system, as shown in Fig. \ref{fig:003_surround_view_maps}. Furthermore, access to the CAN bus interface also provides us with vehicle state and dynamics, including pedal and steering information as shown in Fig. \ref{fig:003_can}.

\section{Sensor Calibration} \label{calibration}
The hardware mounting and calibration of most sensors such as the LIDAR, radar, Mobileye and the GPS/IMU was accomplished by AutonomouStuff, in close collaboration with the manufacturers. The ibeo LUX system includes a fusion box computer which outputs fused data from all LIDAR sensors, providing point clouds (with object classification and tracking) in the vehicle coordinate system $(x,y,z)$, which has its origin at the center of the two rear wheels. The RADAR sensors are also calibrated to output data in the same coordinate system.

\subsection{Time Synchronization}
To synchronize the wide array of sensors equipped in the vehicle, the PolySync bus publishes a synchronized timestamp to all computers equipped in the car used to handle data recording. The published timestamp is based on the IEEE 1588 Precision Time Protocal (PTP), which allows synchronization of time between nodes over an Ethernet network \cite{ieee2002standard}. All data recorded by the testbed have a corresponding timestamp which can be used to achieve easy synchronization.

\subsection{Camera Calibration}
A full surround camera system like ours provides us with a complete view of area around the ego vehicle. But this comes at a considerable overhead during calibration. As far as we are aware, most other vehicular test beds make use of at most one stereo rig, with maybe an additional camera or two. Consequently, there is little reference available when it comes to calibrating a complex setup such as ours.

We based our camera calibration setup on the work of Bo Li et al. \cite{li2013multiple} which proposes the simultaneous calibration (intrinsic and extrinsic) of a system of multiple cameras, where the only assumption is that images from adjacent cameras have a reasonable region of overlap. The camera placement shown in Fig. \ref{fig:002_sensor_layout} along with the focal length of the lens are selected such that there is enough overlap between each pair of neighboring cameras to fit a checkerboard. To start off, the intrinsic parameters (the ones that govern how any point in 3D is projected onto an image) of each camera are estimated using the standard procedure. We use the catadioptric camera model for this step as it is better represents the working of wide angle lenses. Next, we estimate the extrinsic parameters (that represent the relative pose of each camera with respect to one another) between all cameras in a common co-ordinate frame centered about the front camera center. This is done by first constructing a pose graph, where each node of the graph denotes the pose of either a camera, or the calibration pattern relative to a camera at a given instant. The edges between two nodes indicate the capture of the calibration pattern in a specific pose, by a camera at one instant. If a pattern node is connected to two camera nodes, it just indicates that both cameras captured the pattern in a specific pose at the same instant. Finally, initial estimates to the pose of all 8 cameras is obtained by finding the minimal spanning tree and traversing it from top to bottom. With these initial estimates, the poses are refined by non-linear optimization of the re-projection errors in a bundle adjustment like method.

\begin{figure}[!t]
  \centering
  \includegraphics[width=1\linewidth]{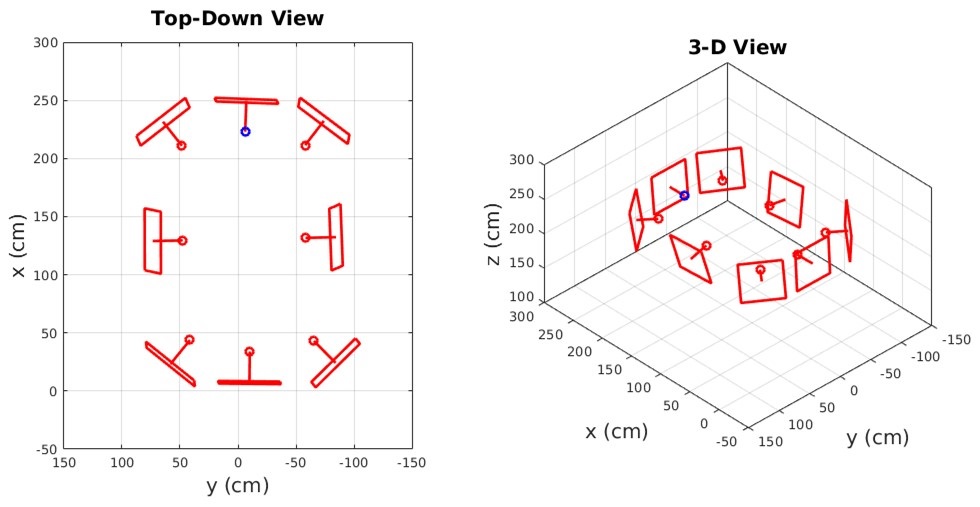}
  \caption{\textbf{Extrinsic camera calibration} Top-down view (left) and 3-D view (right) of the camera pose obtained after calibration.}
  \vspace{-5mm}
  \label{fig:004_camera_calibration}
\end{figure}

As alluded to earlier, a full surround capture system although incredibly useful, results in a
considerable overhead during calibration. This overhead is realized in terms of the actual time it
takes for the calibration to complete, as well as sensitivity of the calibration to small errors. To
ensure that the calibration is highly accurate and reasonably fast to run, we additionally made
the following changes to the calibration process–
\begin{itemize}
\item Replace the feature-based calibration pattern with the standard checkerboard pattern. This improves the accuracy with which the patterns are detected under very high illumination, thereby enabling calibration outdoors. Checkerboard pattern can also detected with sub-pixel accuracy, which results in a more accurate calibration
\item The calibration board is made of glass to ensure near-perfect flatness and rigidity. Glass sheets also do not expand when placed in sunlight for long.
\item The calibration toolbox is set up so that it can parse through videos directly and get frames based on nearest timestamps. This makes it easy while moving the calibration pattern around the vehicle, as only a video needs to be recorded instead of capturing many still images. Also, the videos are read in a multi-threaded fashion which leads to massive speed-ups in the calibration time. All these changes make it possible to perform calibration for each drive if necessary.

\end{itemize}
The final calibration can be verified by viewing the relative pose of cameras with respect to one another as shown in Fig. \ref{fig:004_camera_calibration}. This is done by calculating the baselines (distance between the camera centers) and comparing it to the physically measured values. This however may not yield exact comparisons owing to the fact that the exact location of camera centers is unknown.

\subsection{LIDAR to Camera Calibration}
\begin{figure*}[!t]
  \centering
  \begin{tabular}{c}
        \includegraphics[width =0.95\textwidth]{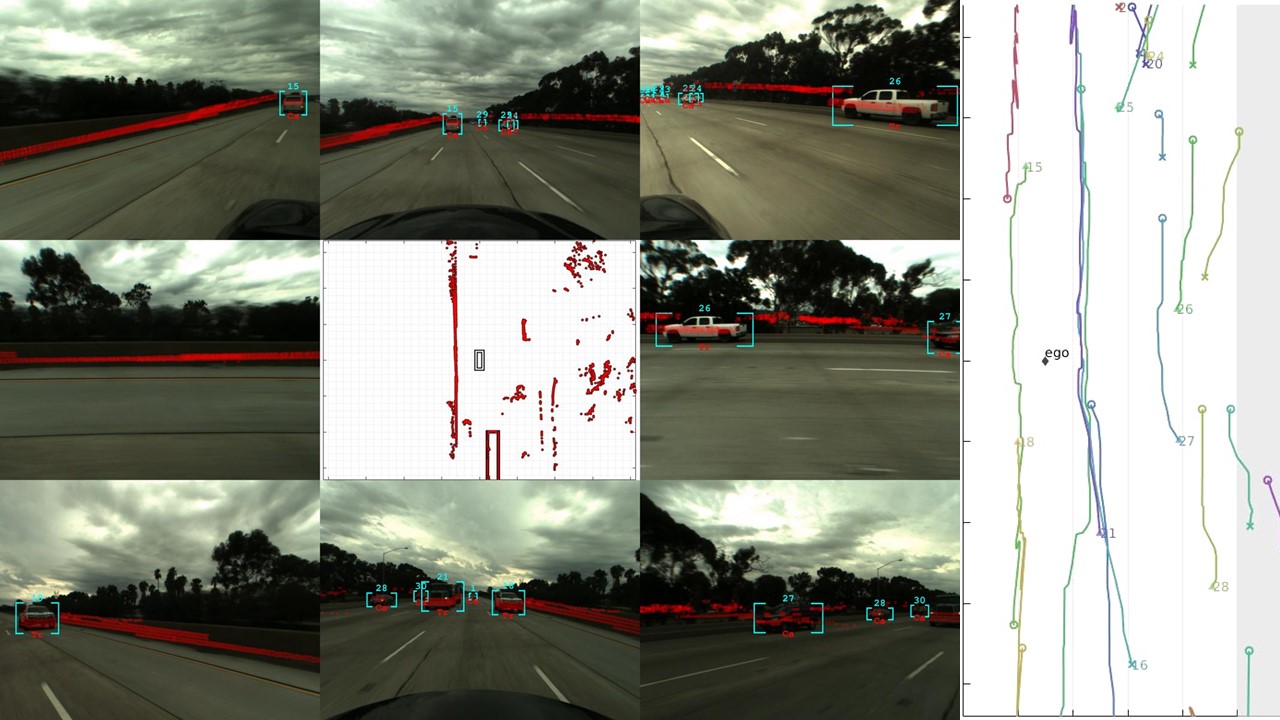}
  \end{tabular}
  \caption{\textbf{LIDAR calibration}. LIDAR points (red-colored dots) are projected onto the 8 camera views after calibration. Sample detections from a vision-based vehicle detector are shown (in cyan) with track IDs listed on top of each detection. The top-down view of surround vehicle trajectories up to the current instant are shown with respect to the ego-vehicle on the right.}
  \vspace{-5mm}
  \label{fig:004_lidar_calibration}
\end{figure*}

After successfully calibrating the cameras, both the LIDAR and camera systems are now capable of localizing objects in 3D in their own co-ordinate systems respectively. However, to make the most off multi-sensor data streams, they need to be calibrated with each other i.e. they must operate in the same frame of reference. This is achieved by finding a transformation (rotation + translation) to take any 3D point in the camera co-ordinate frame to the LIDAR/vehicle coordinate frame. We describe the procedure for LIDAR-Camera calibration below:
\begin{itemize}
\item We placed a flat checkerboard pattern in front the vehicle. The checkerboard pattern is detected by the camera system and the locations of all checkerboard corners in the camera co-ordinate system is obtained. 
\item Then, we find the corresponding points on the checkerboard plane from the LIDAR point clouds. This gives us points on the checkerboard pattern in the LIDAR/vehicle co-ordinate system. 
\item Since the LIDAR points are sparse, it is hard to get point-to-point correspondences between the camera and vehicle co-ordinate frames. Hence, we solve for the transformation in a least squares sense, enforcing the constraint that all points must be \textit{coplanar}. 
\item To do this, we first calculate the normal to the checkerboard plane in both camera and vehicle co-ordinate frames. This is used to estimate the rotation that aligns both sets of surface normals. 
\item Following this, the rotation is held fixed and the translation is calculated in a least squares fashion, by enforcing the condition that all transformed points must lie on the same plane. 
\item Finally, the initial estimates for rotation and translation are optimized by minimizing a non-linear error objective.
\end{itemize}
The LIDAR point clouds can now be projected onto the 8 surround camera views as shown in Fig. \ref{fig:004_lidar_calibration}. Not a lot of vertical information is provided since there are only four horizontal scan layers, although in practice this is rarely necessary. As can be seen, there is a rich amount of detail represented by the red-colored LIDAR points overlayed on each individual image, especially along the horizontal plane.

\section{Concluding Remarks \& Future Work} \label{conclusion}
In this paper, we described our testbed vehicle equipped with a wide range of sensors, constructed with an aim to promote the development of fusion based algorithms in the field. We briefly described the sensor specifications, layout, and outputs. In addition to this, details about sensor calibration and synchronization were provided. We believe that such a heavily sensorized testbed, capable of full surround perception would offer an ideal platform to promote new research in the following areas - 

\begin{itemize}
\item \textit{Vehicle detection and tracking}: The availability of full surround camera data opens up great avenues for fusion based detection algorithms, and multi-target tracking. Information from active sensors could help reduce the false positives of camera based detection systems, and vision based algorithms could help detect farther or smaller objects. Additionally, multi-target tracking could be carried out directly in the real world instead of the image plane, which is considerably more valuable.

\item \textit{Ego-vehicle localization}: Traditional localization algorithms make use of a LIDAR or a vision based sensor in addition to an accurate GPS unit. A testbed such as ours enables researchers to parse full surround information from both cameras and LIDARs along with information about the vehicle dynamics to create better algorithms for this task.

\item \textit{Maneuver Analysis}: With a panoramic view of the world, maneuvers that are longer in duration or are more complex in nature can be captured and analyzed in a more holistic manner. This also motivates researchers to move away from synthetic or simulation based data, and focus on developing algorithms for the real world.

\item \textit{Risk Estimation}: A sensor heavy testbed such as ours ensures that researchers have all the information necessary from the ego vehicles' environment to make a more informed estimate of the risk associated with it. This could also result in the development of proxy visualizations (such as \cite{sivaraman2014dynamic}) that assist the driver in making optimal decisions to ensure safety.

\item \textit{Looking at Humans}: As noted in \cite{ohn2016looking} - ``For a safe
and comfortable ride, intelligent vehicles must observe, understand, model, infer, and predict behavior of occupants inside the vehicle cabin, pedestrians around the vehicle, and humans in surrounding vehicles". The proposed testbed offers multi-sensory data streams to \textit{look} at humans around the vehicles, and in surrounding vehicles.

\end{itemize}

This list is by no means exhaustive, and other novel applications would arise as researchers become more familiar with working on such data.
\section*{ACKNOWLEDGMENT}

We would like to thank Toyota Collaborative Safety Research Center (CSRC) for their generous and continued support. We also express our gratitude towards our colleagues Borhan, Larry, Sourabh, Grady and Nachiket for their assistance in data collection and maintenance of the testbed.


\bibliographystyle{IEEEtran}
\bibliography{bare_jrnl_bib}

\end{document}